\documentclass[10pt,twocolumn,letterpaper]{article}

\usepackage[accsupp]{axessibility}

 \usepackage{cvpr}

\usepackage{multirow}
\usepackage{tikz}
\usetikzlibrary{arrows.meta,positioning,fit}

\usepackage{graphicx}
\usepackage{subcaption}
\usepackage{array}

\definecolor{cvprblue}{rgb}{0.21,0.49,0.74}
\usepackage[pagebackref,breaklinks,colorlinks,allcolors=cvprblue]{hyperref}

\def\paperID{*****}
\def\confName{CVPR}
\def\confYear{2026}

\title{A Deep Learning Iterative Framework for Sentinel-1 Stripmap Enhancement Based on Azimuth Doppler Decomposition}

\author{Juan Francisco Amieva, Christian Ayala\\
Tracasa Instrumental S.L.\\
{\tt\small \{jfamieva,cayala\}@itracasa.es}
\and
Roberto Del Prete\\
European Space Agency\\
{\tt\small roberto.delprete@esa.int}
\and
Mikel Galar\\
Public University of Navarre\\
{\tt\small mikel.galar@unavarra.es}
}

\begin{document}
\maketitle
\begin{abstract}
Synthetic Aperture Radar (SAR) imagery enables all-weather, day-and-night Earth observation; however, it remains difficult to interpret due to speckle noise and other intrinsic imaging artifacts. Sentinel-1 (S1) constitutes one of the most widely used spaceborne SAR missions, offering systematic global coverage, high temporal resolution, dual-polarization imaging, and free data availability. Among S1 modes, Stripmap (SM) provides the highest resolution, yet speckle noise and spatial constraints often hinder applications requiring finer spatial detail.  This motivates the need for effective image enhancement strategies. In this work, we propose a self-supervised enhancement framework for S1 SM imagery based on azimuth subaperture decomposition. The method exploits the physical consistency between subaperture reconstructions and the corresponding full-aperture image to generate paired training data without external sensors, simulated ground truth, or multi-temporal stacks. The proposed framework integrates single- and multi-frame learning and incorporates an iterative inference scheme that progressively refines image quality. Experiments on real S1 SM data show that the proposed approach consistently outperforms the widely adopted self-supervised deep learning baseline MERLIN, in terms of PSNR and SSIM, while MERLIN attains higher ENL, highlighting a trade-off between structural fidelity and speckle smoothing. Overall, the results demonstrate that subaperture-based supervision provides a physically grounded, reproducible, and operationally viable approach for SAR image enhancement using S1 data. It is worth noting that the proposed approach can be extended to other SAR platforms, polarizations, and acquisition modes.
\end{abstract}

\section{Introduction}
\label{sec:intro}
Synthetic Aperture Radar (SAR) has become a cornerstone of operational Earth observation due to its ability to acquire imagery independent of daylight and weather conditions~\cite{torres_2012}. However, SAR data is inherently degraded by speckle noise, which arises from the coherent summation of echoes from elementary scatterers within a resolution cell~\cite{lee_1994}. Furthermore, the side-looking acquisition geometry induces significant radiometric variability driven by local incidence angles and terrain slope~\cite{vollrath_2020}. These effects complicate visual interpretation and can degrade the performance of downstream applications~\cite{zhu_2021,dalsasso_2022,guha_2022}.

Among open-access SAR missions, Sentinel-1 (S1) provides globally consistent C-band acquisitions supporting a broad range of applications \cite{torres_2012}. S1 operates in multiple acquisition modes, with Stripmap (SM) offering the finest spatial resolution (approximately $5\,\mathrm{m}\times5\,\mathrm{m}$ in range $\times$ azimuth) at a narrow swath width of 80 km \cite{esa_2013}. Despite its comparatively higher resolution, SM imagery remains strongly affected by speckle, and its spatial detail is often insufficient to delineate small or densely packed man-made structures~\cite{chini_2018,ayala_2021}. Consequently, recent work has focused on SAR image enhancement techniques that aim to remain consistent with the underlying sensor physics~\cite{zhu_2021,dalsasso_2022}.

Most recent S1‑enhancement studies rely on supervised learning with proxy references, either by using cross‑sensor supervision from high‑resolution commercial SAR data~\cite{wang_2018,ao_2018,amieva_2025} or by supervising across different S1 acquisition modes~\cite{amieva_2026}. While effective, these strategies suffer from practical bottlenecks: they require additional acquisitions, precise co-registration, and often rely on proprietary or limited-access data. In parallel, semi- and self-supervised deep learning approaches such as SAR2SAR~\cite{dalsasso_2021_sar2sar} and MERLIN~\cite{dalsasso_2022} have demonstrated that speckle reduction can be learned directly from SAR data without clean ground truth. However, these methods are tailored specifically for despeckling and do not explicitly address broader enhancement, where noise suppression and spatial resolution improvement are considered jointly.

Azimuth subaperture decomposition is a well-established SAR processing technique that partitions the full azimuth Doppler spectrum into discrete sub-bands and reconstructs an image from each via inverse transformation~\cite{an_2025}. Restricting the azimuth bandwidth coarsens the effective azimuth resolution while preserving the original acquisition geometry and underlying scattering mechanisms. Consequently, the resulting subaperture images exhibit reduced effective spatial resolution and distinct speckle realizations compared to the full-aperture image, even though the ground-sampling-distance (GSD) remains unchanged.

Recent learning-based approaches have exploited these properties in different ways. Wang \etal~\cite{wang_2023} use subaperture images as auxiliary representations for guiding complex-valued neural networks in SAR target recognition, leveraging the fact that each subaperture emphasizes different Doppler components of the same scene. Similarly, Ristea \etal~\cite{ristea_2022} employ azimuth subaperture decomposition to generate multi-view training data for self-supervised representation learning. This strategy effectively creates diverse views of a scene without altering the spatial support or requiring external annotations.

In the context of image restoration, An~\etal~\cite{an_2025} utilize subaperture decomposition for SAR despeckling by treating subaperture images as multiple observations of an identical scene generated under the same imaging mechanism. Their formulation demonstrates that subaperture data can effectively suppress speckle while remaining consistent with SAR signal physics. Conversely, Dong~\etal~\cite{dong_2025} integrate subaperture decomposition into a deep learning framework for complex-valued super-resolution (SR), leveraging multi-scale sub-band information to enhance azimuth detail reconstruction. However, the literature currently treats these tasks separately: subapertures are employed either for despeckling or for resolution enhancement. To date, no approach has explicitly leveraged this physics-grounded property for constructing self-supervised training pairs to jointly address speckle reduction and spatial detail recovery.

To address these limitations, we propose a framework that leverages azimuth subaperture decomposition to construct self-supervised input--target pairs directly from S1 SM acquisitions. By exploiting the physical consistency between azimuth-band-limited subaperture reconstructions and the corresponding full-aperture image, the proposed approach enables sensor-consistent enhancement without requiring external reference sensors, simulated ground truth, or multi-temporal stacks. The proposed framework could also be applied to other sensors, acquisition modes, and polarizations.

This paper makes the following contributions:
\begin{itemize}

    \item We introduce a self-supervised data generation strategy for S1 SM imagery based on azimuth subaperture decomposition, enabling physics-consistent training without external supervision.

    \item We investigate single- and multi-frame SAR enhancement within a unified learning framework.

    \item We propose an iterative inference scheme that exposes the trade-off between speckle suppression and structural detail preservation.

\end{itemize}

\section{Related Work}
\label{sec:related}

\subsection{Speckle Reduction for SAR Images}
\label{sec:rw_despeckling}

\textbf{Classical Statistical and Nonlocal Methods.} Early despeckling approaches relied on local statistics and simplified speckle models to design adaptive smoothing filters, such as the Lee and Kuan filters~\cite{lee_1983,kuan_1985}. These methods exploit spatial correlations via sliding-window estimators. However, while effective at speckle suppression, they often degrade edge sharpness and fine textures. The introduction of Non-Local Means~\cite{buades_2005} established the value of nonlocal patch similarity, leading to probabilistic frameworks~\cite{deledalle_2009} and collaborative filtering methods like SAR-BM3D~\cite{parrilli_2012}. Despite their effectiveness in preserving fine details, these classical methods are often limited by high computational complexity and strong parameter dependence~\cite{perera_2022}.

\noindent\textbf{Deep Learning-based Despeckling.} Recent despeckling literature is dominated by deep learning, ranging from supervised to self-supervised approaches~\cite{fracastoro_2021,zhu_2021}. Supervised methods typically depend on pseudo-ground truth obtained via multi-temporal averaging, multi-looking, or algorithmic reconstruction~\cite{chierchia_2017,lattari_2019,dalsasso_2020,mullissa_2020}. Since ``clean'' reference images are rarely available in real-world SAR scenarios, recent research has pivoted toward self-supervised learning by adapting principles from natural image denoising methods like Noise2Noise and blind-spot learning~\cite{lehtinen_2018,krull_2019}. These approaches include SAR2SAR~\cite{dalsasso_2021_sar2sar}, MERLIN~\cite{dalsasso_2022}, and blind-spot despeckling in the SAR domain (Speckle2Void)~\cite{molini_2022}, alongside Bernoulli-sampling variants~\cite{yuan_2021}. While these methods mitigate the need for paired data, they hinge on assumptions, such as temporal consistency in multi-image settings and spatial noise independence in blind-spot formulations, that may be violated in dynamic scenes or structured targets. This limitation motivates the integration of physics-aware priors into learning frameworks~\cite{fracastoro_2021}.

\subsection{Learning-based Spatial Resolution Enhancement in SAR}
\label{sec:rw_sr}

\noindent\textbf{Super-Resolution in Computer Vision.} Traditional image SR methods, such as bicubic interpolation, are limited in their ability to reconstruct sharp edge structures~\cite{massarelli_2026}. Over the past decade, deep learning-based single-image~SR~(SISR) approaches have gained significant attention by learning a mapping between low- and high-resolution images~\cite{yang_2019}. Early architectures like SRCNN~\cite{dong_2015} applied shallow convolutional networks to pre-upsampled inputs, focusing on refining high-frequency details rather than explicitly increasing sampling distance within the network itself. Subsequent models, such as EDSR~\cite{lim_2017}, significantly improved reconstruction accuracy through deeper residual networks and learnable upsampling. More recently, attention mechanisms have pushed the state of the art by modeling long-range spatial dependencies, resulting in enhanced detail recovery. Prominent examples include RCAN~\cite{zhang_2018}, SAN~\cite{dai_2019}, and ESRT~\cite{lu_2022}.

Beyond single-image techniques, Multi-Frame SR (MFSR) exploits complementary information across multiple frames to improve reconstruction. CNN-based methods like EDVR~\cite{wang_2019} utilize spatiotemporal attention for feature fusion, while recurrent frameworks such as BasicVSR++~\cite{chan_2022} leverage bidirectional propagation to capture long-term context. Recent Video Restoration Transformers (e.g., VRT, RVRT) further enhance this by using self-attention to model spatiotemporal features explicitly~\cite{liang_2022,liang_2024}.

\noindent\textbf{SAR Super-Resolution and Cross-Sensor Supervision.} SAR SR has primarily relied on GAN-based translation and reconstruction approaches, such as S1-to-TerraSAR-X mapping~\cite{ao_2018} and various SR-specific architectures~\cite{wang_2018,zheng_2019}. Some studies also explored optical-guided strategies to inject spatial details from co-registered optical imagery~\cite{li_2022}. While visually appealing, cross-sensor or cross-modality supervision suffers from domain gaps such as differing carrier frequencies, incidence angles, processing chains, and scene-dependent scattering which complicate validation and risk introducing modality-specific artifacts inconsistent with SAR physics~\cite{zhu_2021}. Furthermore, most existing approaches primarily focus on spatial resolution enhancement and do not explicitly address speckle suppression, which remains intrinsic to coherent SAR imaging. To bridge the gap between data-driven models and physical constraints, hybrid formulations such as compressed sensing and algorithm-unrolling have been explored~\cite{guha_2022}. However, the consensus in recent literature suggests that SAR SR requires supervision that is both sensor-consistent and faithful to SAR physics, prioritizing reconstruction fidelity over photorealistic synthesis~\cite{blau_2018,zhu_2021}.

Amieva~\etal\cite{amieva_2026} has addressed this challenge by proposing a joint SR and despeckling framework for S1 Interferometric Wide Swath (IW) imagery, utilizing S1 SM as a reference. While this strategy maintained sensor consistency, the cross-mode supervision (IW$\rightarrow$SM) still introduced inevitable domain shifts that limited the fidelity of the reconstruction.

\subsection{Physics-grounded Supervision via Azimuth Subapertures}
\label{sec:rw_subaperture}

\textbf{Azimuth Subaperture Decomposition and Effective Resolution.} In SAR imaging, spatial resolution is determined by different physical mechanisms in the range and azimuth dimensions. The slant-range resolution is primarily controlled by the transmitted bandwidth $B$~\cite{dong_2025}, approximated as:
\begin{equation}
\rho_r \approx \frac{c}{2B},
\label{eq:range_res}
\end{equation}
where $c$ is the speed of light. Conversely, the azimuth \mbox{resolution} is intrinsically linked to the Doppler bandwidth $B_D$ accumulated over the synthetic aperture. For a platform velocity $v$ and physical antenna length $L$, this relationship is given by:
\begin{equation}
B_D \approx \frac{2v}{L},
\qquad
\rho_a \approx \frac{v}{B_D} = \frac{L}{2},
\label{eq:az_res_doppler}
\end{equation}

Azimuth subaperture decomposition partitions the full azimuth Doppler spectrum into $K$ distinct sub-bands and reconstructs an image from each, obtaining subaperture looks that represent the same scene with reduced effective azimuth resolution~\cite{dong_2025,an_2025} (See \cref{fig:subap_decomp_process}). Specifically, if the $k$-th subaperture retains a fraction $\alpha_k = B_{D,k}/B_D$ of the total Doppler bandwidth, its effective azimuth resolution approximately degrades to:
\begin{equation}
\rho_{a,k} \approx \frac{\rho_a}{\alpha_k},
\label{eq:subap_res}
\end{equation}

This degradation does not modify the spatial sampling of the final ground-projected images since identical geocoding and resampling operators are applied. Both subaperture and full-aperture products can therefore share the same ground-range grid. The difference appears as reduced high-Doppler support and altered speckle statistics rather than spatial decimation~\cite{an_2025}.

Consequently, learning a mapping from subaperture inputs to full-aperture references can be formulated as a \textit{same-grid} enhancement problem, which does not require SR-specific upsampling modules. In this setting, fully convolutional encoder-decoder architectures (e.g., U-Net variants) are particularly well-suited, as they naturally process fixed-size inputs while capturing multi-scale contextual features necessary for detail recovery.

\noindent\textbf{Subaperture-aware Learning.} Recent literature has begun to exploit subaperture decomposition as a source of complementary views for representation learning and downstream SAR understanding tasks~\cite{ristea_2022, ristea_2023}. In high-level tasks, subaperture cues have been utilized for Doppler-driven ship detection and velocity estimation~\cite{iqbal_2025}, while early explorations into subaperture-based supervision for target recognition were conducted by Wang \etal~\cite{wang_2023}. In the context of restoration, subapertures have been employed to denoise full-aperture imagery while mitigating the resolution loss associated with classical multi-looking~\cite{an_2025}, and into fusion strategies for complex-valued SR frameworks~\cite{dong_2025}. Unlike these approaches, our work leverages azimuth subaperture reconstruction as a scalable, physics-grounded mechanism to synthesize paired LR/HR supervision on the same grid. This strategy is specifically tailored for S1 SM enhancement, prioritizing conservative detail recovery and speckle reduction consistent with the sensor's native imaging physics.

\section{Methodology}
\label{sec:method}

\subsection{Dataset}
\label{sec:dataset}

We developed a self-supervised dataset using S1 SM Single Look Complex (SLC) dual-polarization (VV + VH) data. The SM acquisition mode was selected due to its continuous azimuth illumination and larger intrinsic azimuth bandwidth compared to the standard IW mode, enabling higher azimuth resolution. However, SM acquisitions are less frequent and typically driven by specific tasking requests~\cite{esa_2013}.

Overall, the dataset consists of 10 acquisitions selected from the limited available archive to ensure geographical diversity across continental regions, covering a total area of 162,305.4~km$^2$. To ensure geometric consistency and maximize data availability, we focused exclusively on ascending orbits, leaving the integration of descending passes for future research.
\vspace{-6pt}
\paragraph{Pre-processing.} All SLC products were processed using the Sentinel Application Platform (SNAP)~\cite{Snap}. Specifically, we performed precise orbit correction, radiometric calibration, and terrain correction, adhering to the standard preprocessing workflow for S1 data following the procedures outlined by Filipponi~\etal~\cite{filipponi_2019}. The resulting calibrated and georeferenced backscatter intensities were converted to a decibel (dB) scale to transform multiplicative speckle into approximately additive noise \cite{deledalle_2012}.  To ensure training stability across heterogeneous scenes, we mitigated the impact of outliers by clipping the intensity values for each polarization channel at the $0.1^{th}$ and $99.9^{th}$ percentiles computed over the entire dataset. Finally, the clipped values were linearly rescaled to the $[0,1]$ range to facilitate gradient-based optimization.
\vspace{-6pt}
\paragraph{Splitting Protocol and Patch Extraction.}

To mitigate spatial leakage caused by high correlation between neighboring patches, we partitioned the dataset at the acquisition level. In this regard, entire scenes were assigned to either the training, validation, or test sets, ensuring that the model is evaluated on geographically unseen regions. For training and validation, we extracted non-overlapping $96\times96$ patches from each acquisition, discarding any tiles containing no-data values. Conversely, testing is performed on full scenes using a sliding window with 50\% overlap. Final dataset statistics are provided in \cref{tab:dataset_stats}.

\begin{table}[!h]
\caption{Dataset extension statistics.}
\label{tab:dataset_stats}
\centering
\small
\begin{tabular}{lccc}
\toprule
Split & \# Acquisitions & Area (km$^2$) & \# Patches \\
\midrule
Train & ${5}$ & $84,401.3$ & $301,573$ \\
Validation  & ${3}$ & $49,434.5$ & $210,467$ \\
Test  & ${2}$ & $28,469.7$ & -- \\
\bottomrule
\end{tabular}
\end{table}

\subsection{Subapertures Generation}
\label{sec:subaps_gen}

\begin{figure}[!h]
    \centering
    \includegraphics[width=0.9\linewidth]{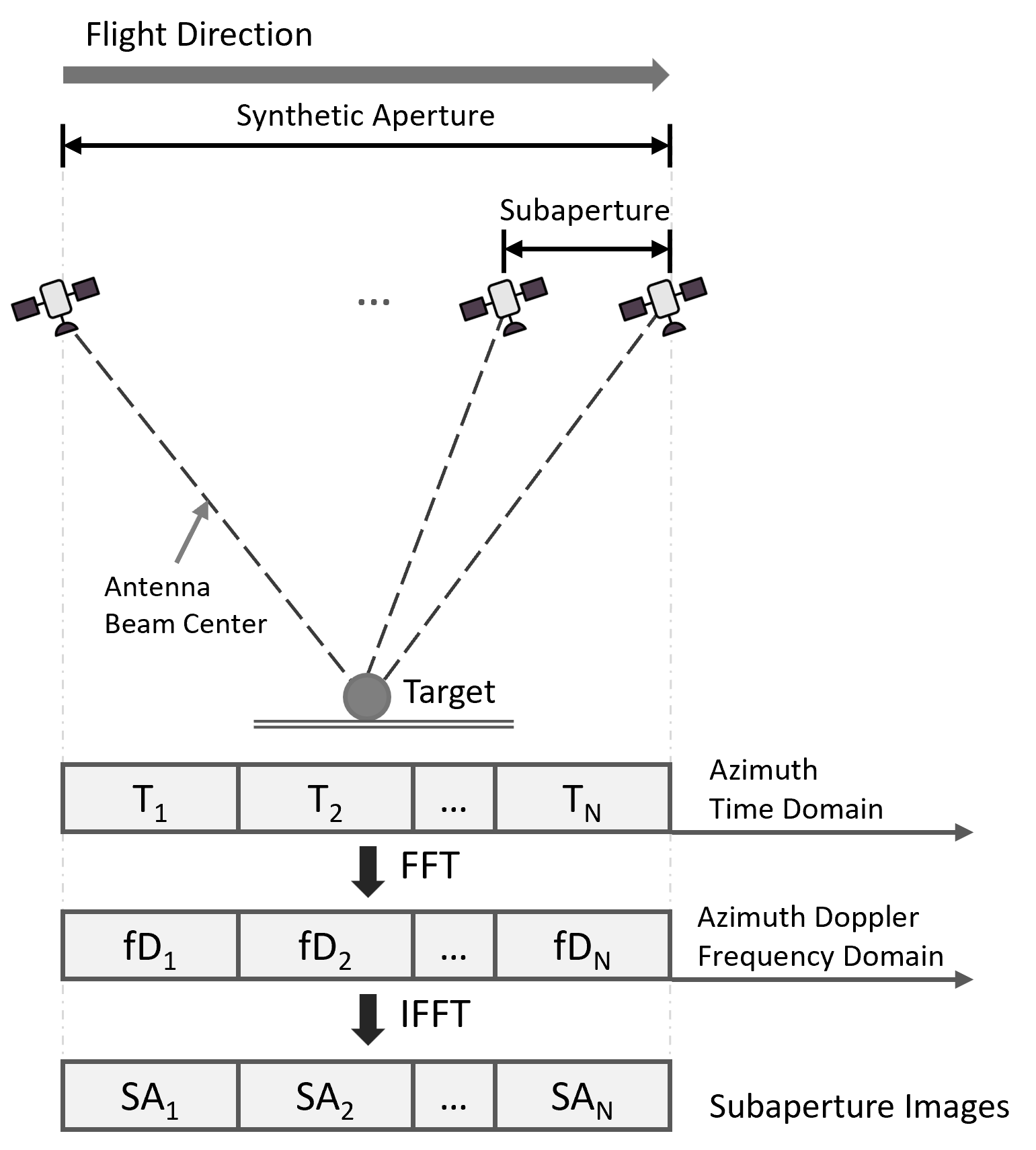}
    \caption{Simplified schema of subaperture decomposition process.}
    \label{fig:subap_decomp_process}
\end{figure}

Training inputs are generated from SLC products via azimuth subaperture decomposition.
Sub-band boundaries are derived from product metadata by constructing the azimuth frequency grid and evenly partitioning the azimuth bandwidth into three non-overlapping intervals. The complex SLC data is transformed into the azimuth Doppler frequency domain via a 1-D Fast Fourier Transform (FFT). We then perform spectrum normalization and a de-weighting step to compensate for the azimuth apodization (Hamming-type weighting). Each interval is then extracted, re-centered in the spectral domain, and transformed back to the spatial domain via an Inverse FFT (IFFT)~\cite{an_2025}, yielding one complex-valued look per interval. Finally, a circular shift is applied along the azimuth direction to spatially re-center the reconstructed looks. \cref{fig:subap_decomp_process} presents a simplified schema of the described processing.

Although the azimuth bandwidth could be divided into an arbitrary number of intervals, increasing the number of subapertures reduces the effective Doppler support of each look and consequently degrades azimuth resolution. For this reason, we adopt a three-look, non-overlapping partition of the azimuth Doppler spectrum per acquisition. The choice of this configuration was based on preliminary experiments.

Following decomposition, the subaperture products undergo the same pre-processing workflow described in \cref{sec:dataset}.

\subsection{Problem Setup}
\label{sec:archs}

\begin{figure}[!h]
    \centering
    \includegraphics[width=\linewidth]{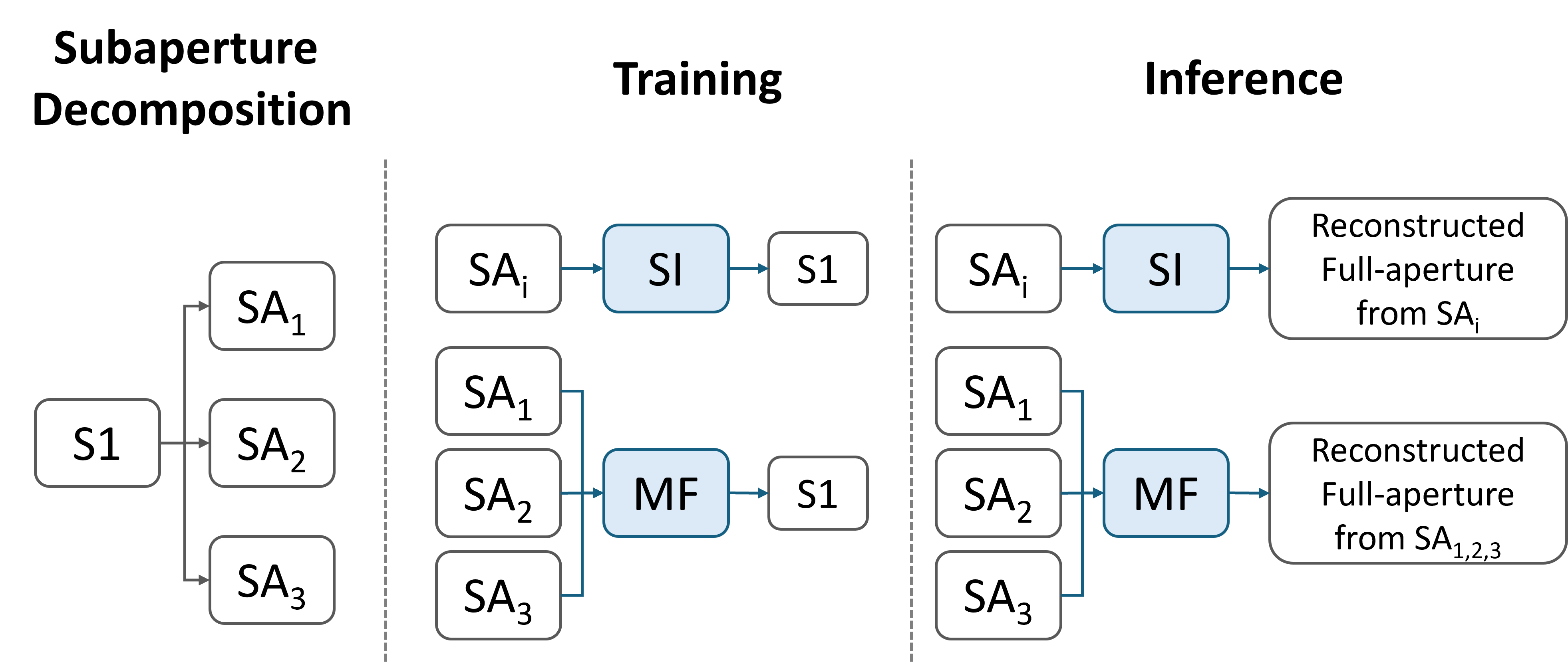}
    \caption{Summarized Problem Setup.}
    \label{fig:problem_setup}
\end{figure}

\paragraph{Training.}
We consider subapertures as inputs and the corresponding full-aperture intensity as reference. We evaluate two distinct configurations, as summarized in \cref{fig:problem_setup}:

\begin{itemize}
    \item \emph{Single-input (SI):} This configuration takes a single subaperture patch as input to predict the full-aperture reference.  We employ \emph{MA-Net}~\cite{fan_2020}, a standard U-Net-like architecture with multi-scale attention, as the backbone for pixel-wise intensity regression.
    \item \emph{Multi-frame (MF):} This setup jointly processes all three subaperture patches to predict the full-aperture reference. We adopt a commonly used multi-frame fusion architecture, \emph{FPN-ConvLSTM}~\cite{chamorro_2021}, which integrates multi-scale feature extraction via a Feature Pyramid Network (FPN) with ConvLSTM layers to fuse information across the subaperture sequence before decoding the final intensity prediction.
\end{itemize}

\vspace{-6pt}
\paragraph{Inference.} Given the distinct architectural designs of the two frameworks, the inference strategy varies slightly between them. For \emph{SI}, inference is performed independently on each subaperture, resulting in three separate reconstructions that can be analyzed individually or averaged. In contrast, \emph{MF} processes the three subapertures jointly to produce a single reconstruction (see \cref{fig:problem_setup}). In operational settings, the SI configuration processes a single full-aperture S1 SM acquisition, whereas the MF configuration jointly processes three co-registered S1 SM acquisitions.

\subsection{Iterative Refinement}
\label{sec:iter_infer}

\begin{figure}[!h]
    \centering
    \includegraphics[width=\linewidth]{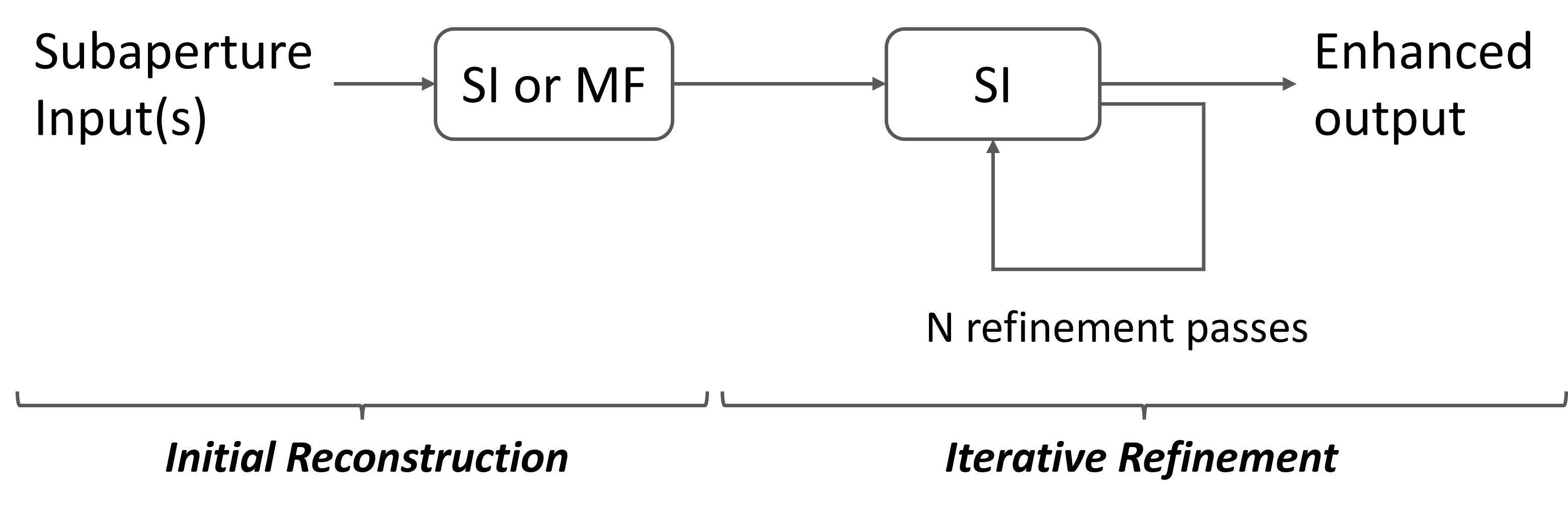}
    \caption{Iterative Refinement Process.}
    \label{fig:iterative_refinement_process}
\end{figure}

Given that both proposed frameworks produce an enhanced full-aperture reconstruction from synthesized subapertures, we investigate the potential for further enhancement gains through iterative refinement. Specifically, we evaluate a recursive inference scheme where the enhanced output from an initial pass is treated as a ``pseudo-subaperture" and fed back into the \emph{SI} architecture. While the \emph{MF} model requires a triplet of distinct subapertures and thus cannot be directly iterated, its initial output can serve as the starting point for subsequent refinement steps using the \emph{SI} model. This approach allows us to explore whether the learned denoising and reconstruction priors can be recursively applied to progressively refine the enhanced output. \cref{fig:iterative_refinement_process} illustrates this iterative procedure.

\subsection{Experimental Setup}
\label{sec:exp_setup}

Both configurations are trained using the Adam optimizer~\cite{kingma_2015} for 100 epochs, using a OneCycle learning rate scheduler~\cite{smith_2018} with a maximum learning rate of $10^{-3}$. Due to hardware memory constraints, \emph{SI} is trained with a batch size of 512, while \emph{MF} utilizes a batch size of 32. All experiments are conducted on an NVIDIA RTX A5500 GPU (24 GB VRAM).

The training objective compares the network prediction $\hat{y}$ with the full-aperture reference $y$ using a weighted combination of an $\ell_2$ fidelity, Structural Similarity Index (SSIM), and a Kernel Density Estimation (KDE)-based distribution matching term~\cite{gomez_alanis_2020}:
\vspace{-2pt}
\begin{equation}
\mathcal{L} = \alpha\,\|\hat{y}-y\|_2^2 + \beta\,(1-\mathrm{SSIM}(\hat{y},y)) + \gamma\,\mathcal{L}_{\mathrm{KDE}}(\hat{y},y).
\label{eq:loss_setup}
\end{equation}

where $\alpha$, $\beta$, and $\gamma$ weight the $\ell_2$, SSIM, and KDE terms, respectively. We set $\alpha=0.2$, $\beta=0.3$, and $\gamma=0.5$ based on preliminary experiments.

Data augmentation is limited to dihedral transformations (combination of rotations by multiples of $90^\circ$ and flips), applied consistently to both inputs and targets. To improve robustness against the Doppler partitioning scheme, \emph{SI} randomly samples one of the three subapertures as input during each training step, while \emph{MF} is trained with random permutations of the input subaperture order. Furthermore, we apply a patch-wise histogram matching step to align the backscatter intensity distributions of the subaperture inputs with the full-aperture reference. The purpose of this operation is to ensure that the input distributions are consistent with those used during the model’s operational phase, which relies on full-aperture input images.

\subsection{Evaluation Framework}
\label{sec:eval_frame}

We evaluate our models using Peak Signal-to-Noise Ratio (PSNR) and SSIM~\cite{wang_2004} to assess reconstruction fidelity against the full-aperture reference. Since the full-aperture reference is affected by speckle, no noise-free ground truth is available; thus, PSNR and SSIM measure consistency rather than absolute fidelity.
To quantify despeckling performance, we utilize the Equivalent Number of Looks (ENL)~\cite{ko_2022}, averaged over 20 manually selected homogeneous ROIs within the test set. It is worth mentioning that these ROIs have been manually selected to ensure no structural features were present. All metrics are reported as averages across the test acquisitions.

Metrics are computed according to the inference configuration:
\begin{itemize}
    \item \emph{SI}: Inference is performed independently on each of the three subapertures; metrics are calculated per output and subsequently averaged.
    \item \emph{MF}: The three subapertures are processed jointly to generate a single reconstruction, with metrics computed directly on this output.
    \item \emph{Iterative Inference:} Metrics are computed at every refinement step, according to the corresponding input for the iterative process (\emph{SI} or \emph{MF}).
\end{itemize}

Finally, we compare our approach against \emph{MERLIN}~\cite{dalsasso_2022} as a baseline using two distinct strategies. First, we apply \emph{MERLIN} to each subaperture independently and average the results, following the \emph{SI} protocol as a Full-Reference assessment. Second, to provide a broader comparison, we \mbox{apply} \emph{MERLIN} directly to the full-aperture images. Since no noise-free ground truth exists for the full-aperture data, this constitutes a No-Reference analysis; consequently, we only report the ENL to evaluate the degree of speckle reduction without calculating fidelity metrics. Qualitative visual comparisons are additionally provided.

\section{Results and Discussion}
\label{sec:results}

Following the framework described in \cref{sec:eval_frame}, this section addresses the following research questions:
\vspace{6pt}
\begin{enumerate}
    \item \emph{Comparative Evaluation of SI, MF, and MERLIN}: How do the proposed \emph{SI} and \emph{MF} configurations compare in terms of reconstruction fidelity and despeckling, and how do they perform relative to the \emph{MERLIN} baseline?

    \item \emph{Iterative Refinement}: What is the impact of recursive \emph{SI} inference, and how does the trade-off between speckle reduction and structural detail preservation evolve across multiple refinement passes?

    \item \emph{Qualitative Assessment in an Operational Scenario}: Are the learned enhancement characteristics preserved on real full-aperture S1 SM imagery?
\end{enumerate}
\vspace{6pt}

The following sections present the corresponding analyses to address each research question.

\subsection{Comparative Evaluation of SI, MF, and MERLIN}
\label{sec:res_q1}

\cref{tab:q1_main} reports the quantitative results for the proposed \emph{SI} and \emph{MF} configurations, alongside the \emph{MERLIN} baseline, in terms of SSIM, PSNR, and ENL metrics computed for both polarizations (VV and VH). Note that the best results for each metric are highlighted in \textbf{boldface}.

\begin{table}[!h]
\centering
\caption{\emph{SI} and \emph{MF} compared with \emph{MERLIN}.}
\label{tab:q1_main}
\small
\begin{tabular*}{\linewidth}{@{\extracolsep{\fill}} l rr rr rr@{}}
\toprule
& \multicolumn{2}{c}{SSIM $\uparrow$} & \multicolumn{2}{c}{PSNR $\uparrow$} & \multicolumn{2}{c}{ENL $\uparrow$} \\
\cmidrule(lr){2-3} \cmidrule(lr){4-5} \cmidrule(l{3pt}){6-7}
Method & VV & VH & VV & VH & VV & VH \\
\midrule
SI                      & 70.5 & 62.4 & 29.2 & 27.6 & 28  & 26 \\
MF                      & \textbf{84.2} & \textbf{78.4} & \textbf{30.3} & \textbf{28.2} & 14  & 13 \\
MERLIN$_{\text{Sub}}$   & 63.5 & 45.4 & 19.6 & 12.2 & 66  & 50 \\
MERLIN$_{\text{Full}}$  & --   & --   & --   & --   & \textbf{294} & \textbf{81} \\
\bottomrule
\end{tabular*}
\end{table}

\begin{figure*}[!h]
    \centering
    \scriptsize
    \addtolength{\tabcolsep}{-5pt}
    \renewcommand{\arraystretch}{1}
    \begin{tabular}{cccccc}
       \includegraphics[width=0.167\textwidth]{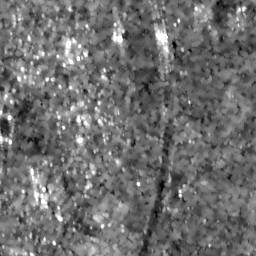} &
        \includegraphics[width=0.167\textwidth]{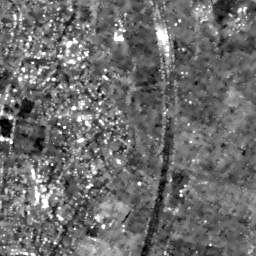} &
        \includegraphics[width=0.167\textwidth]{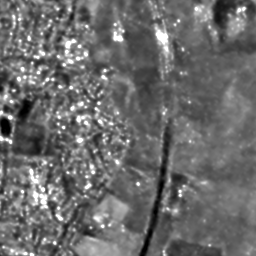} &
      \includegraphics[width=0.167\textwidth]{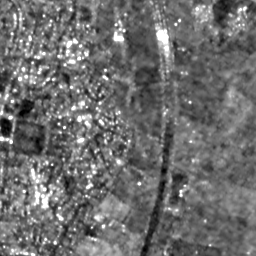} &
        \includegraphics[width=0.167\textwidth]{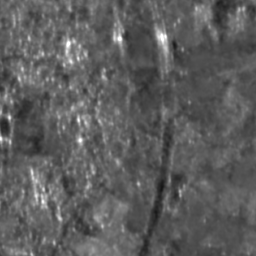} &
        \includegraphics[width=0.167\textwidth]{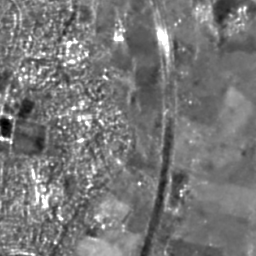}  \\
       SA$_1$ & Full Aperture Reference & \emph{SI} & \emph{MF}  & $\emph{MERLIN}_{\text{Sub}}$ & $\emph{MERLIN}_{\text{Full}}$  \\
       & ( PSNR /  SSIM)  & ( 26.80 dB /  69.81) & ( 31.24 dB /  86.59) & ( 25.22 dB /  59.28) & \vspace{4pt} \\

  \end{tabular}\vspace{-5pt}
  \caption{Visual comparison (VV polarization) between the \emph{MERLIN} baseline and the proposed \emph{SI} and \emph{MF} methods. Only the first subaperture (SA$_1$) is shown due to space constraints.}
  \label{fig:q1_res}
\end{figure*}

Overall, both \emph{SI} and \emph{MF} substantially outperform the \emph{MERLIN} baseline applied to subaperture inputs ($\emph{MERLIN}_{\text{Sub}}$), yielding significant gains in reconstruction fidelity across both polarizations. The \emph{MF} configuration achieves the highest performance, reaching a PSNR of $30.3/28.2$~dB and an SSIM of $0.842/0.784$. While the \emph{SI} model yields lower fidelity metrics, it achieves a higher ENL than \emph{MF}, suggesting more aggressive speckle suppression. In contrast, $\emph{MERLIN}_{\text{Full}}$ produces markedly higher ENL values (294/81), reflecting substantial smoothing when applied directly to full-aperture imagery. However, this increase in ENL is not necessarily accompanied by improved structural fidelity. \cref{fig:q1_res} reinforces these quantitative findings through a qualitative comparison, highlighting the trade-off between noise reduction and structural preservation across the different methods.

\subsection{Iterative Refinement}
\label{sec:res_q2}

This section investigates the impact of recursive \emph{SI} inference and the resulting trade-off between speckle reduction and structural preservation. Quantitative results for varying refinement passes, using initial reconstructions generated by either \emph{SI} or \emph{MF}, are summarized in \cref{tab:si_refinement}.

\begin{table}[!h]
\centering
\caption{Effect of additional \emph{SI} refinement passes. ``Init.'' denotes the model used to generate the initial reconstruction (\emph{SI} or \emph{MF}). ``Passes'' denotes the number of additional \emph{SI} refinement passes.}
\label{tab:si_refinement}
\small
\setlength{\tabcolsep}{3.5pt}
\begin{tabular*}{\linewidth}{@{\extracolsep{\fill}}c r rr rr rr@{}}
\toprule
\multirow{2}{*}{Init.} & \multirow{2}{*}{Passes} &
\multicolumn{2}{c}{SSIM $\uparrow$} &
\multicolumn{2}{c}{PSNR $\uparrow$} &
\multicolumn{2}{c}{ENL $\uparrow$} \\
\cmidrule(l{3pt}r{1pt}){3-4}\cmidrule(l{2pt}r{1pt}){5-6}\cmidrule(l{2pt}){7-8}
& & VV & VH & VV & VH & VV & VH \\

\midrule
\multirow{5}{*}{SI}
& - & 70.5 & 62.4 & 29.2 & 27.6 & 28  & 26  \\
& 1 & 68.8 & 60.3 & 27.8 & 26.3 & 52  & 55  \\
& 2 & 66.2 & 57.1 & 26.3 & 24.7 & 105 & 102 \\
& 3 & 63.6 & 53.9 & 25.0 & 23.3 & 201 & 207 \\
& 4 & 61.4 & 51.2 & 24.0 & 22.2 & \textbf{274} & \textbf{395} \\
\midrule
\multirow{5}{*}{MF}
& - & \textbf{84.2} & \textbf{78.4} & \textbf{30.3} & \textbf{28.2} & 14  & 13  \\
& 1 & 78.9 & 73.1 & 29.2 & 27.8 & 31  & 31  \\
& 2 & 72.8 & 66.2 & 27.1 & 26.0 & 60  & 68  \\
& 3 & 67.8 & 60.3 & 25.4 & 24.3 & 113 & 124 \\
& 4 & 63.9 & 55.5 & 24.1 & 22.9 & 213 & 235 \\
\bottomrule
\end{tabular*}
\vspace{-2mm}
\end{table}

Across both polarizations, increasing the number of refinement passes results in a consistent increase in ENL and a corresponding decrease in PSNR and SSIM. This trend explicitly quantifies the speckle--detail trade-off: while repeated passes enhance image smoothness, they progressively degrade reconstruction fidelity. For instance, when the initial reconstruction is produced by \emph{SI}, two additional passes boost ENL from $28/26$ to $105/102$ (VV/VH), yet PSNR drops from $29.2/27.6$~dB to $26.3/24.7$~dB, and SSIM from $0.705/0.624$ to $0.662/0.571$. At these high pass counts, ENL values approach or exceed the \emph{MERLIN}$_{\emph{Full}}$ benchmark; however, the substantial degradation in fidelity indicates a increased risk of over-smoothing and loss of critical spatial features.

A particularly favorable operating point emerges when applying a single \emph{SI} refinement pass to the \emph{MF} output. This configuration effectively shifts the high-fidelity \emph{MF} result toward stronger despeckling. Specifically, one SI pass on the \emph{MF} initial reconstruction increases ENL from $14/13$ to $31/31$, while maintaining a robust SSIM of $0.789/0.731$.

\begin{figure*}[!h]
    \centering
    \scriptsize
    \addtolength{\tabcolsep}{-5pt}
    \renewcommand{\arraystretch}{1}
    \begin{tabular}{ccccc}
       \includegraphics[width=0.19\textwidth]{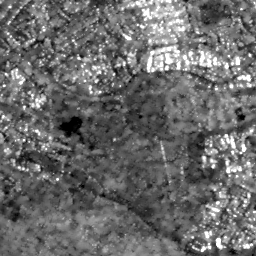} &
        \includegraphics[width=0.19\textwidth]{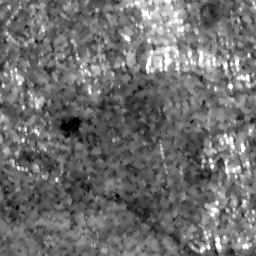} &
        \includegraphics[width=0.19\textwidth]{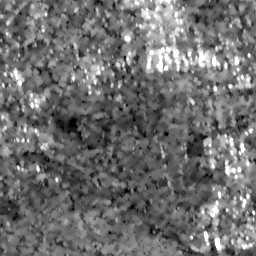} &
        \includegraphics[width=0.19\textwidth]{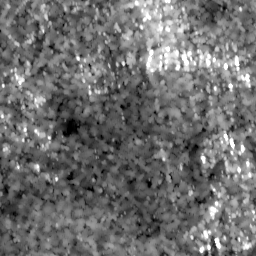} &
           \\
       Full Aperture Reference & SA$_1$ &  SA$_2$ &  SA$_3$  &  \\
       ( PSNR /  SSIM) &  &  &  &  \\

       \includegraphics[width=0.19\textwidth]{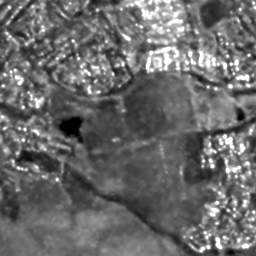} &
        \includegraphics[width=0.19\textwidth]{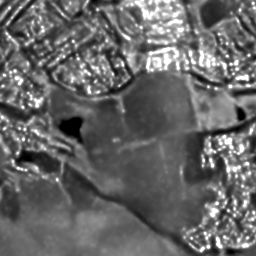} &
        \includegraphics[width=0.19\textwidth]{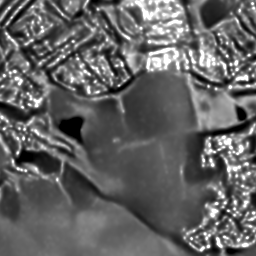} &
        \includegraphics[width=0.19\textwidth]{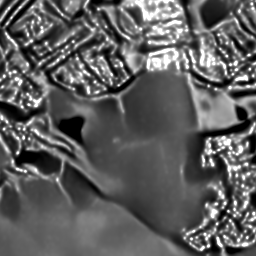} &
        \includegraphics[width=0.19\textwidth]{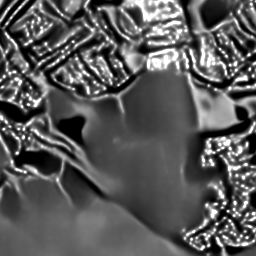}  \\
       \emph{SI} & $\emph{SI}_{\text{1}}$ & $\emph{SI}_{\text{2}}$ & $\emph{SI}_{\text{3}}$  & $\emph{SI}_{\text{4}}$\\
       ( 26.52 dB / 71.01) & ( 25.10 dB /  66.25)  & ( 23.29 dB /  59.77) & ( 21.52 dB /  52.86) & ( 19.98 dB /  46.60)  \\

       \includegraphics[width=0.19\textwidth]{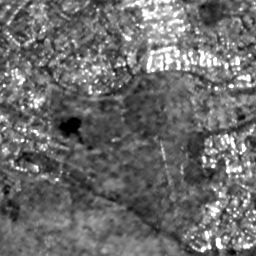} &
        \includegraphics[width=0.19\textwidth]{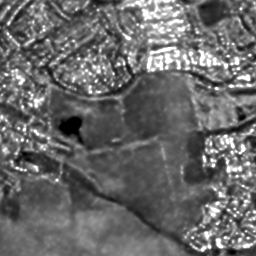} &
        \includegraphics[width=0.19\textwidth]{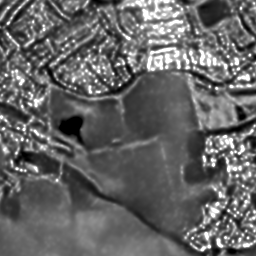} &
        \includegraphics[width=0.19\textwidth]{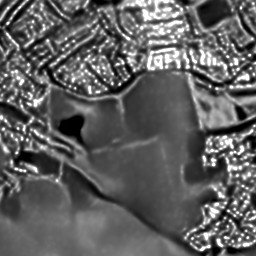} &
        \includegraphics[width=0.19\textwidth]{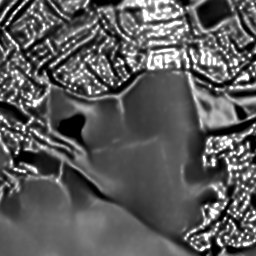}  \\
       \emph{MF} & $\emph{MF}_{\text{1}}$ & $\emph{MF}_{\text{2}}$ & $\emph{MF}_{\text{3}}$  & $\emph{MF}_{\text{4}}$\\
       ( 29.37 dB /  81.84) & ( 27.69 dB /  77.03)  & ( 25.27 dB /  69.51) & ( 22.94 dB /  61.13) & ( 21.02 dB /  53.23)  \\

  \end{tabular}\vspace{-5pt}
  \caption{Qualitative comparison (VV polarization) of subaperture inputs, full-aperture reference, and reconstructions. The second and third rows correspond to \emph{SI} and \emph{MF} models; subindices denote inference refinement passes (1--4).}
  \label{fig:q2_res}
\end{figure*}

Overall, applying \emph{SI} refinement passes to the \emph{MF} output yields a more balanced reconstruction than the \emph{SI} model alone. The \emph{MF} formulation provides complementary information across subaperture inputs, establishing a strong baseline that subsequent \emph{SI} refinement further regularizes. Qualitative examples in \cref{fig:q2_res} support this, showing that the first few refinement passes significantly reduce speckle with limited structural alterations, while excessive iterations eventually compromise fine details.

\subsection{Qualitative Assessment in an Operational Scenario}
\label{sec:res_q3}

\begin{figure*}[!h]
    \centering
    \scriptsize
    \addtolength{\tabcolsep}{-5pt}
    \renewcommand{\arraystretch}{1}
    \begin{tabular}{cccccc}
       \includegraphics[width=0.3\textwidth]{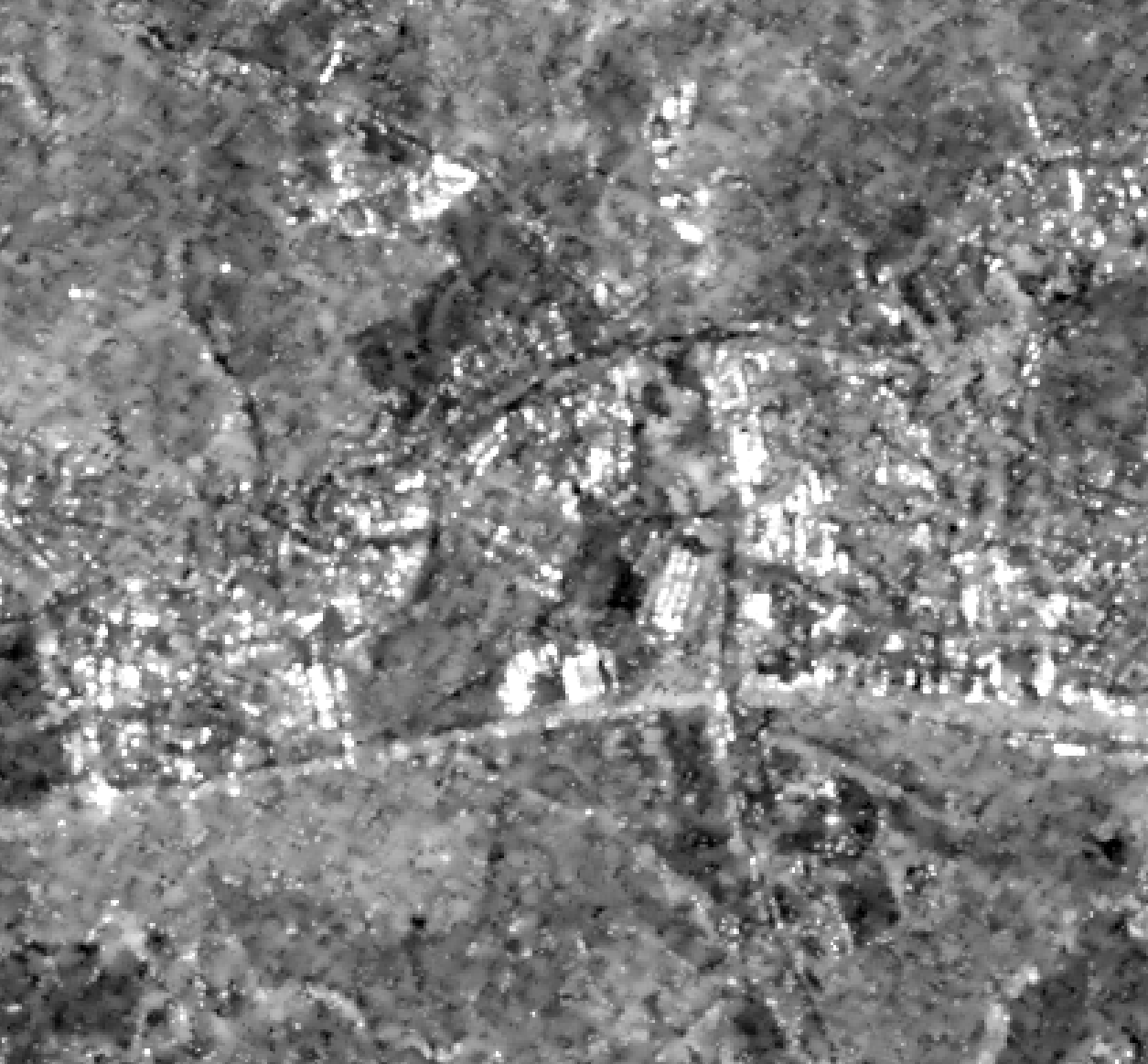} &
        \includegraphics[width=0.3\textwidth]{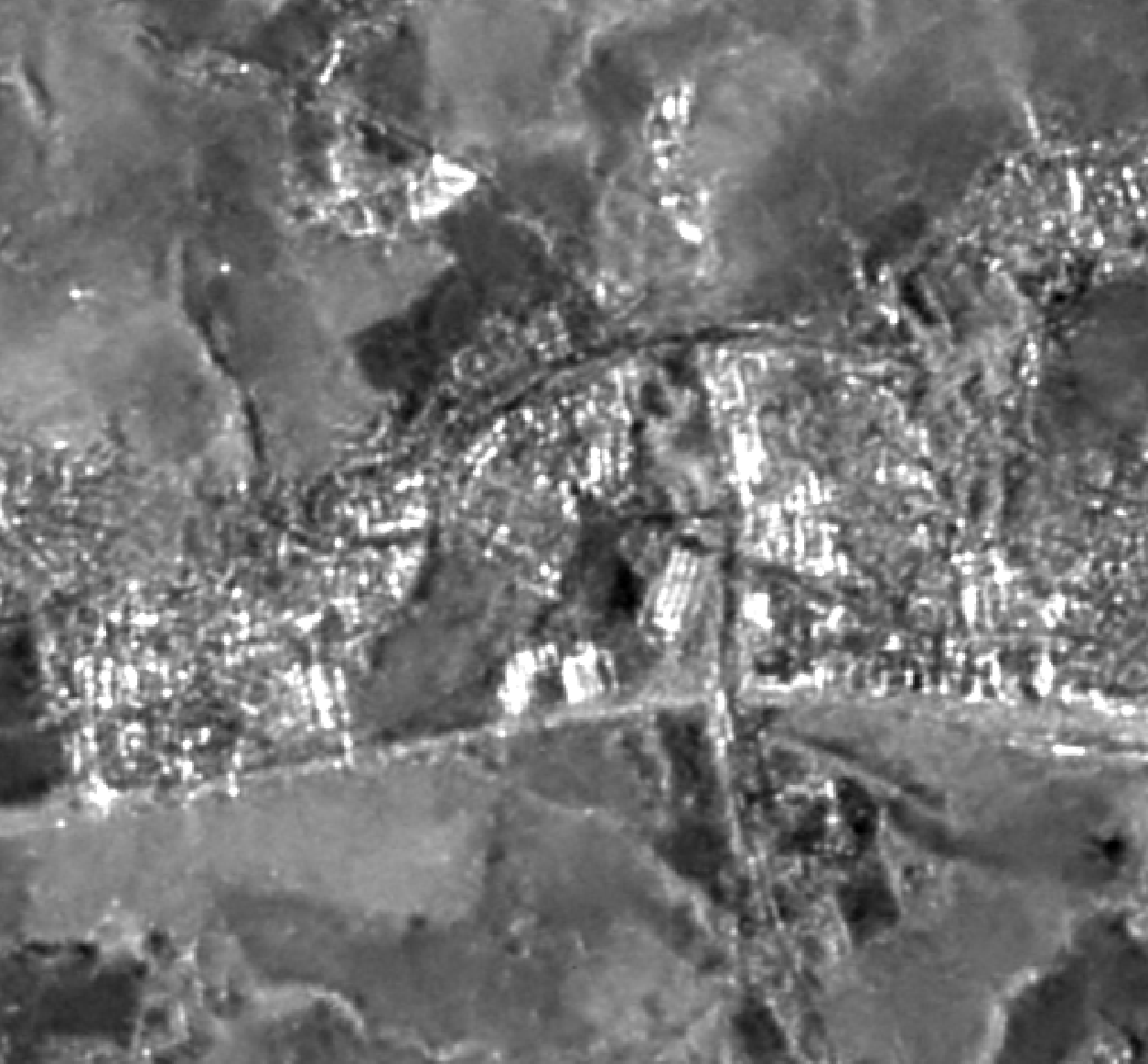} &
        \includegraphics[width=0.3\textwidth]{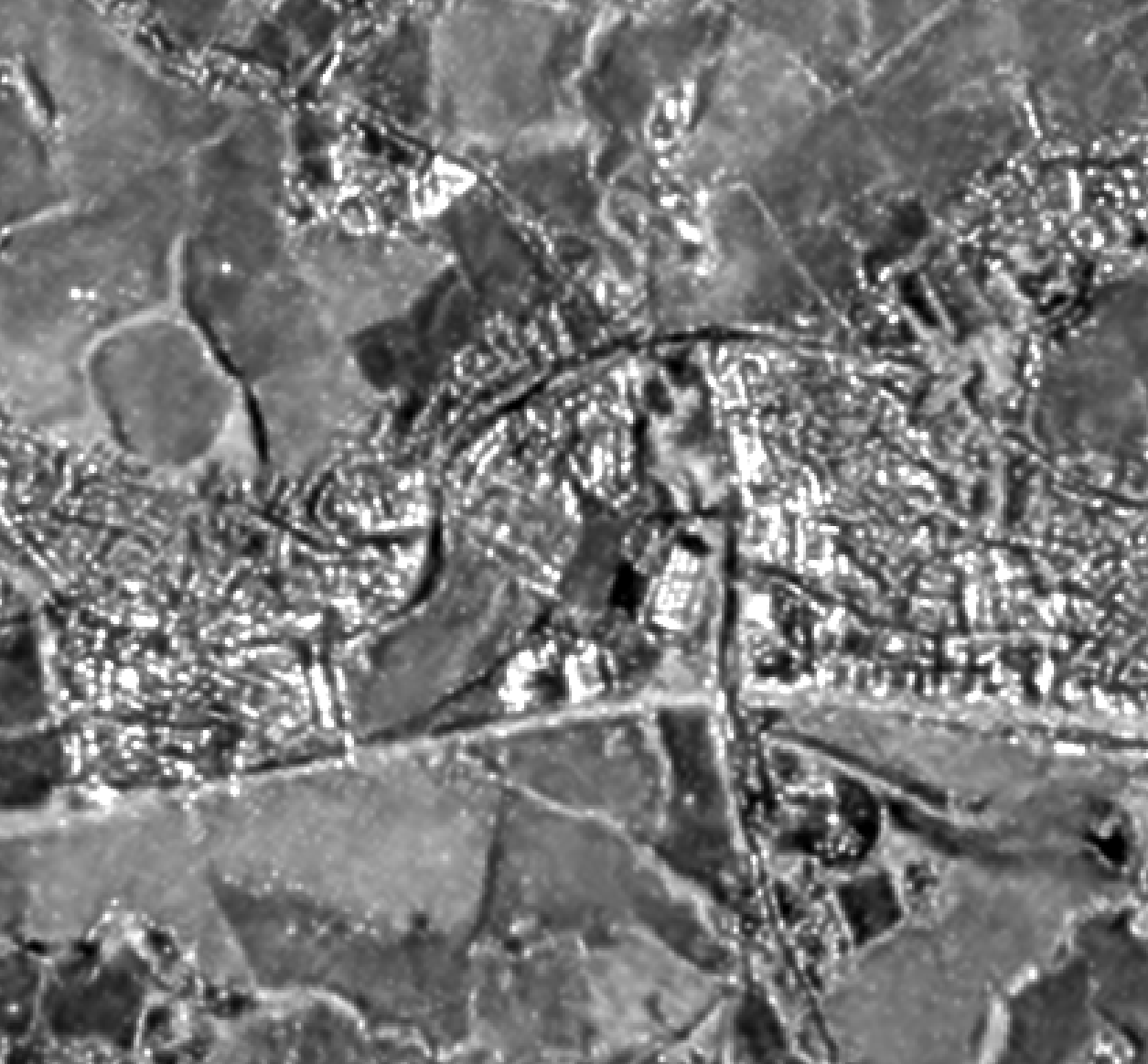} &\\
       Full Aperture Input (S1 SM) & $\emph{MERLIN}_{\text{Full}}$ & \emph{$MF_1$} \\

  \end{tabular}\vspace{-5pt}

    \caption{Visual results on real full-aperture S1 SM imagery (VV polarization), comparing the proposed $\emph{MF}_1$ versus \emph{MERLIN}$_{\text{Full}}$.}
\label{fig:q3_res}
\end{figure*}

Since no reference data is available for full-aperture inputs, quantitative evaluation cannot be performed in this operational setting.
Therefore, \cref{fig:q3_res} presents qualitative results obtained by applying \emph{MERLIN} and our best-performing configuration \emph{$MF_1$} directly to full-aperture S1 SM imagery. For a fair comparison with \emph{MERLIN}, the same S1 SM acquisition was replicated three times to match the multi-frame structure of \emph{MF}, preventing the exploitation of multi-temporal information.

The proposed \emph{$MF_1$} exhibits improved structural sharpness while avoiding the stronger smoothing observed in $\emph{MERLIN}_{\text{Full}}$.

Moreover, to complement the presented results and better assess achievable reconstruction quality, full-acquisition inference examples are provided as supplementary material at \href{https://huggingface.co/datasets/juanfra54/Ai4space_inference}{[url]}.

\section{Conclusions and Future Work}
\label{sec:conclusion}

In this work, we have presented a self-supervised framework for enhancing S1 SM imagery by leveraging azimuth Doppler subaperture decomposition. By treating the full-aperture intensity as a sensor-consistent reference, we developed two distinct architectures: Single-Input (SI) and Multi-Frame (MF), that effectively learn to reconstruct high-fidelity images while suppressing speckle noise.

Our results demonstrate that the \emph{MF} configuration, which exploits diversity across the azimuth Doppler spectrum, achieves higher structural consistency with the full-aperture reference. Conversely, the \emph{SI} model provides stronger despeckling capabilities. We further explored iterative refinement, showing that recursive passes through the \emph{SI} architecture substantially increase ENL but eventually lead to diminishing returns, where high-frequency details are compromised. Finally, compared to the baseline self-supervised enhancement method \mbox{\emph{MERLIN}}, the proposed approach achieves superior overall performance.

For future work, we aim to extend this framework to other sensors, acquisition modes, polarizations, and flight directions, as well as to incorporate multi-temporal sequences to further improve the robustness of the self-supervised priors. We also plan to investigate the effect of varying the number of azimuth subapertures used for supervision, analyzing how this design choice influences reconstruction fidelity and speckle suppression. Furthermore, we will assess the impact of these enhanced products on downstream tasks such as maritime object detection and land-cover classification, assessing whether the improved visual quality translates into higher automated analysis performance. Finally, we will explore alternative strategies for constructing more robust reference targets to enable a more comprehensive quantitative evaluation of enhancement performance.

\section{Acknowledgements}
This work was also supported by project PID2022-136627NB-I00 (MCIN/AEI/10.13039/501100011033 FEDER, EU).

{
    \small
    \bibliographystyle{ieeenatfullname}
    \bibliography{main}

@String(CVPR= {IEEE Conf. Comput. Vis. Pattern Recog.})

@String(ECCV= {Eur. Conf. Comput. Vis.})

@String(ICLR = {Int. Conf. Learn. Represent.})

@String(CVPR  = {CVPR})

@String(ECCV  = {ECCV})

@String(ICLR  = {ICLR})

@inproceedings{torres_2012,
	title        = {The Sentinel-1 mission and its application capabilities},
	author       = {Torres, Ramon and Snoeij, Paul and Davidson, Malcolm and Bibby, David and Lokas, Steve and others},
	year         = 2012,
	booktitle    = {2012 IEEE International Geoscience and Remote Sensing Symposium (IGARSS)},
	publisher    = {IEEE},
	pages        = {1703--1706}
}

@article{ao_2018,
	title        = {Dialectical GAN for SAR image translation: From Sentinel-1 to TerraSAR-X},
	author       = {Ao, Dongyang and Dumitru, Corneliu Octavian and Schwarz, Gottfried and Datcu, Mihai},
	year         = 2018,
	journal      = {Remote Sensing},
	publisher    = {MDPI},
	volume       = 10,
	number       = 10,
	pages        = 1597
}

@article{vollrath_2020,
	title        = {Angular-Based Radiometric Slope Correction for Sentinel-1 on Google Earth Engine},
	author       = {Vollrath, Andreas and Mullissa, Adugna and Reiche, Johannes},
	year         = 2020,
	journal      = {Remote Sensing},
	volume       = 12,
	number       = 11,
	article-number = 1867
}

@article{lee_1994,
	title        = {Speckle filtering of synthetic aperture radar images: A review},
	author       = {Jong-Sen Lee and L. Jurkevich and Piet Dewaele and Patrick Wambacq and Andr{\'e} Oosterlinck},
	year         = 1994,
	journal      = {Remote Sensing Reviews},
	volume       = 8,
	number       = 4,
	pages        = {313--340}
}

@article{lee_1983,
	title        = {Digital image smoothing and the sigma filter},
	author       = {Jong-Sen Lee},
	year         = 1983,
	journal      = {Computer Vision, Graphics, and Image Processing},
	volume       = 24,
	number       = 2,
	pages        = {255--269}
}

@article{kuan_1985,
	title        = {Adaptive Noise Smoothing Filter for Images with Signal-Dependent Noise},
	author       = {Kuan, Darwin T. and Sawchuk, Alexander A. and Strand, Timothy C. and Chavel, Pierre},
	year         = 1985,
	journal      = {IEEE Transactions on Pattern Analysis and Machine Intelligence},
	volume       = {PAMI-7},
	number       = 2,
	pages        = {165--177}
}

@inproceedings{buades_2005,
	title        = {A non-local algorithm for image denoising},
	author       = {Buades, Antoni and Coll, Bartomeu and Morel, J-M},
	year         = 2005,
	booktitle    = {2005 IEEE computer society conference on computer vision and pattern recognition (CVPR'05)},
	volume       = 2,
	pages        = {60--65},
	organization = {Ieee}
}

@article{deledalle_2009,
	title        = {Iterative Weighted Maximum Likelihood Denoising With Probabilistic Patch-Based Weights},
	author       = {Charles-Alban Deledalle and Lo{\"i}c Denis and Florence Tupin},
	year         = 2009,
	journal      = {IEEE Transactions on Image Processing},
	volume       = 18,
	number       = 12,
	pages        = {2661--2672}
}

@article{parrilli_2012,
	title        = {A Nonlocal SAR Image Denoising Algorithm Based on LLMMSE Wavelet Shrinkage},
	author       = {Parrilli, Sara and Poderico, Mariana and Angelino, Cesario Vincenzo and Verdoliva, Luisa},
	year         = 2012,
	journal      = {IEEE Transactions on Geoscience and Remote Sensing},
	volume       = 50,
	number       = 2,
	pages        = {606--616}
}

@inproceedings{perera_2022,
	title        = {Transformer-Based SAR Image Despeckling},
	author       = {Perera, Malsha V. and Bandara, Wele Gedara Chaminda and Valanarasu, Jeya Maria Jose and Patel, Vishal M.},
	year         = 2022,
	booktitle    = {IGARSS 2022 - 2022 IEEE International Geoscience and Remote Sensing Symposium},
	volume       = {},
	number       = {},
	pages        = {751--754}
}

@article{fracastoro_2021,
	title        = {Deep learning methods for synthetic aperture radar image despeckling: An overview of trends and perspectives},
	author       = {Fracastoro, Giulia and Magli, Enrico and Poggi, Giovanni and Scarpa, Giuseppe and Valsesia, Diego and Verdoliva, Luisa},
	year         = 2021,
	journal      = {IEEE Geoscience and Remote Sensing Magazine},
	publisher    = {IEEE},
	volume       = 9,
	number       = 2,
	pages        = {29--51}
}

@article{zhu_2021,
	title        = {Deep learning meets SAR: Concepts, models, pitfalls, and perspectives},
	author       = {Zhu, Xiao Xiang and Montazeri, Sina and Ali, Mohsin and Hua, Yuansheng and Wang, Yuanyuan and Mou, Lichao and Shi, Yilei and Xu, Feng and Bamler, Richard},
	year         = 2021,
	journal      = {IEEE Geoscience and Remote Sensing Magazine},
	publisher    = {IEEE},
	volume       = 9,
	number       = 4,
	pages        = {143--172}
}

@inproceedings{chierchia_2017,
	title        = {{SAR} Image Despeckling Through Convolutional Neural Networks},
	author       = {Chierchia, Giovanni and Cozzolino, Davide and Poggi, Giovanni and Scarpa, Giuseppe and Verdoliva, Luisa},
	year         = 2017,
	booktitle    = {2017 IEEE International Geoscience and Remote Sensing Symposium (IGARSS)}
}

@article{lattari_2019,
	title        = {Deep Learning for {SAR} Image Despeckling},
	author       = {Lattari, Federico and Gonz{\'a}lez, Claudio and Le{\'o}n, D. and Asaro, F. and Matteucci, Matteo},
	year         = 2019,
	journal      = {Remote Sensing},
	volume       = 11,
	number       = 13
}

@article{dalsasso_2020,
	title        = {{SAR} Image Despeckling by Deep Neural Networks: From a Pre-Trained Model to an End-to-End Training Strategy},
	author       = {Dalsasso, Emanuele and Yang, Xiangli and Denis, Lo{\"\i}c and Tupin, Florence and Yang, Wen},
	year         = 2020,
	journal      = {Remote Sensing},
	volume       = 12,
	number       = 16
}

@article{mullissa_2020,
	title        = {DeSpeckNet: Generalizing deep learning-based SAR image despeckling},
	author       = {Mullissa, Adugna G and Marcos, Diego and Tuia, Devis and Herold, Martin and Reiche, Johannes},
	year         = 2020,
	journal      = {IEEE Transactions on Geoscience and Remote Sensing},
	publisher    = {IEEE},
	volume       = 60,
	pages        = {1--15}
}

@article{yuan_2021,
	title        = {An Advanced SAR Image Despeckling Method by Bernoulli-Sampling-Based Self-Supervised Deep Learning},
	author       = {Yuan, Ye and Wu, Yanxia and Fu, Yan and Wu, Yulei and Zhang, Lidan and Jiang, Yan},
	year         = 2021,
	journal      = {Remote Sensing},
	volume       = 13,
	number       = 18
}

@article{molini_2022,
	title        = {Speckle2Void: Deep Self-Supervised SAR Despeckling With Blind-Spot Convolutional Neural Networks},
	author       = {Molini, Andrea Bordone and Valsesia, Diego and Fracastoro, Giulia and Magli, Enrico},
	year         = 2022,
	journal      = {IEEE Transactions on Geoscience and Remote Sensing},
	volume       = 60,
	number       = {},
	pages        = {1--17}
}

@article{dalsasso_2021_sar2sar,
	title        = {{SAR2SAR}: A Semi-Supervised Despeckling Algorithm for {SAR} Images},
	author       = {Dalsasso, Emanuele and Denis, Lo{\"\i}c and Tupin, Florence},
	year         = 2021,
	journal      = {IEEE Journal of Selected Topics in Applied Earth Observations and Remote Sensing},
	volume       = 14,
	pages        = {4321--4329}
}

@article{dalsasso_2022,
	title        = {As If by Magic: Self-Supervised Training of Deep Despeckling Networks with {MERLIN}},
	author       = {Dalsasso, Emanuele and Denis, Lo{\"\i}c and Tupin, Florence},
	year         = 2022,
	journal      = {IEEE Transactions on Geoscience and Remote Sensing},
	volume       = 60
}

@article{dong_2015,
	title        = {Image super-resolution using deep convolutional networks},
	author       = {Dong, Chao and Loy, Chen Change and He, Kaiming and Tang, Xiaoou},
	year         = 2015,
	journal      = {IEEE transactions on pattern analysis and machine intelligence},
	publisher    = {IEEE},
	volume       = 38,
	number       = 2,
	pages        = {295--307}
}

@article{yang_2019,
	title        = {Deep Learning for Single Image Super-Resolution: A Brief Review},
	author       = {Yang, Wenming and Zhang, Xuechen and Tian, Yapeng and Wang, Wei and Xue, Jing-Hao and Liao, Qingmin},
	year         = 2019,
	journal      = {IEEE Transactions on Multimedia},
	volume       = 21,
	number       = 12,
	pages        = {3106--3121}
}

@inproceedings{lim_2017,
	title        = {Enhanced deep residual networks for single image super-resolution},
	author       = {Lim, Bee and Son, Sanghyun and Kim, Heewon and Nah, Seungjun and Mu Lee, Kyoung},
	year         = 2017,
	booktitle    = {Proceedings of the IEEE conference on computer vision and pattern recognition workshops},
	pages        = {136--144}
}

@inproceedings{dai_2019,
	title        = {Second-order attention network for single image super-resolution},
	author       = {Dai, Tao and Cai, Jianrui and Zhang, Yongbing and Xia, Shu-Tao and Zhang, Lei},
	year         = 2019,
	booktitle    = {Proceedings of the IEEE/CVF conference on computer vision and pattern recognition},
	pages        = {11065--11074}
}

@inproceedings{zhang_2018,
	title        = {Image super-resolution using very deep residual channel attention networks},
	author       = {Zhang, Yulun and Li, Kunpeng and Li, Kai and Wang, Lichen and Zhong, Bineng and Fu, Yun},
	year         = 2018,
	booktitle    = {Proceedings of the European conference on computer vision (ECCV)},
	pages        = {286--301}
}

@inproceedings{lu_2022,
	title        = {Transformer for single image super-resolution},
	author       = {Lu, Zhisheng and Li, Juncheng and Liu, Hong and Huang, Chaoyan and Zhang, Linlin and Zeng, Tieyong},
	year         = 2022,
	booktitle    = {Proceedings of the IEEE/CVF conference on computer vision and pattern recognition},
	pages        = {457--466}
}

@inproceedings{wang_2019,
	title        = {Edvr: Video restoration with enhanced deformable convolutional networks},
	author       = {Wang, Xintao and Chan, Kelvin CK and Yu, Ke and Dong, Chao and Change Loy, Chen},
	year         = 2019,
	booktitle    = {Proceedings of the IEEE/CVF conference on computer vision and pattern recognition workshops},
	pages        = {0--0}
}

@inproceedings{chan_2022,
	title        = {Basicvsr++: Improving video super-resolution with enhanced propagation and alignment},
	author       = {Chan, Kelvin CK and Zhou, Shangchen and Xu, Xiangyu and Loy, Chen Change},
	year         = 2022,
	booktitle    = {Proceedings of the IEEE/CVF conference on computer vision and pattern recognition},
	pages        = {5972--5981}
}

@article{liang_2024,
	title        = {Vrt: A video restoration transformer},
	author       = {Liang, Jingyun and Cao, Jiezhang and Fan, Yuchen and Zhang, Kai and Ranjan, Rakesh and Li, Yawei and Timofte, Radu and Van Gool, Luc},
	year         = 2024,
	journal      = {IEEE Transactions on Image Processing},
	publisher    = {IEEE},
	volume       = 33,
	pages        = {2171--2182}
}

@article{liang_2022,
	title        = {Recurrent video restoration transformer with guided deformable attention},
	author       = {Liang, Jingyun and Fan, Yuchen and Xiang, Xiaoyu and Ranjan, Rakesh and Ilg, Eddy and Green, Simon and Cao, Jiezhang and Zhang, Kai and Timofte, Radu and Gool, Luc V},
	year         = 2022,
	journal      = {Advances in Neural Information Processing Systems},
	volume       = 35,
	pages        = {378--393}
}

@inproceedings{amieva_2025,
	title        = {Deep Learning for Enhancing Sentinel-1 Imagery},
	author       = {Amieva, Juan Francisco and Ayala, Christian and Galar, Mikel},
	year         = 2025,
	booktitle    = {2025 IEEE International Geoscience and Remote Sensing Symposium (IGARSS)}
}

@article{amieva_2026,
	title        = {A deep learning approach to jointly super-resolve and despeckle Sentinel-1 imagery},
	author       = {Amieva, Juan Francisco and Ayala, Christian and Galar, Mikel},
	year         = 2026,
	journal      = {Acta Astronautica},
	volume       = 238,
	pages        = {1396--1407}
}

@inproceedings{blau_2018,
	title        = {The Perception-Distortion Tradeoff},
	author       = {Blau, Yochai and Michaeli, Tomer},
	year         = 2018,
	booktitle    = {Proceedings of the IEEE/CVF Conference on Computer Vision and Pattern Recognition (CVPR)},
	pages        = {6228--6237}
}

@article{lehtinen_2018,
	title        = {Noise2Noise: Learning image restoration without clean data},
	author       = {Lehtinen, Jaakko and Munkberg, Jacob and Hasselgren, Jon and Laine, Samuli and Karras, Tero and Aittala, Miika and Aila, Timo},
	year         = 2018,
	journal      = {arXiv preprint arXiv:1803.04189}
}

@inproceedings{krull_2019,
	title        = {Noise2void-learning denoising from single noisy images},
	author       = {Krull, Alexander and Buchholz, Tim-Oliver and Jug, Florian},
	year         = 2019,
	booktitle    = {Proceedings of the IEEE/CVF conference on computer vision and pattern recognition},
	pages        = {2129--2137}
}

@inproceedings{wang_2018,
	title        = {Super-resolution SAR Image Reconstruction via Generative Adversarial Network},
	author       = {Wang, Longgang and Zheng, Mana and Du, Wenbo and Wei, Menglin and Li, Lianlin},
	year         = 2018,
	booktitle    = {2018 IEEE International Symposium on Antennas and Propagation and USNC-URSI Radio Science Meeting (ISAPE)}
}

@inproceedings{zheng_2019,
	title        = {Self-Normalizing Generative Adversarial Network for Super-Resolution Reconstruction of SAR Images},
	author       = {Zheng, Ce and Jiang, Xue and Zhang, Ye and Liu, Xingzhao and Yuan, Bin and Li, Zhixin},
	year         = 2019,
	booktitle    = {IGARSS 2019 - 2019 IEEE International Geoscience and Remote Sensing Symposium},
	pages        = {1911--1914}
}

@article{li_2022,
	title        = {OGSRN: Optical-guided super-resolution network for SAR image},
	author       = {Yanshan, LI and Li, ZHOU and Fan, XU and Shifu, CHEN},
	year         = 2022,
	journal      = {Chinese Journal of Aeronautics},
	publisher    = {Elsevier},
	volume       = 35,
	number       = 5,
	pages        = {204--219}
}

@inproceedings{guha_2022,
	title        = {Sar Super-Resolution Using Physics-Aware Adaptive Compressed Sensing},
	author       = {Guha, Sanhita and Datcu, Mihai and Ender, Joachim},
	year         = 2022,
	booktitle    = {IGARSS 2022-2022 IEEE International Geoscience and Remote Sensing Symposium},
	pages        = {52--55},
	organization = {IEEE}
}

@article{ayala_2021,
	title        = {A Deep Learning Approach to an Enhanced Building Footprint and Road Detection in High-Resolution Satellite Imagery},
	author       = {Ayala, Christian and Sesma, Rub{\'e}n and Aranda, Carlos and Galar, Mikel},
	year         = 2021,
	journal      = {Remote Sensing},
	volume       = 13,
	number       = 16,
	pages        = 3135
}

@article{chini_2018,
	title        = {Towards a 20 m Global Building Map from Sentinel-1 {SAR} Data},
	author       = {Chini, Marco and Pelich, Ramona and Hostache, Radu and Matgen, Patrick and Lopez-Martinez, Carlos},
	year         = 2018,
	journal      = {Remote Sensing},
	volume       = 10,
	number       = 11,
	pages        = 1833
}

@misc{esa_2013,
	title        = {Sentinel-1 User Handbook},
	author       = {{European Space Agency (ESA)}},
	year         = 2013,
	howpublished = {ESA Standard Document, Issue 1 Rev 0}
}

@article{dong_2025,
	title        = {Complex-Valued SAR Image Super-Resolution via Subaperture Learning and Fusion},
	author       = {Dong, Ganggang and Wang, Yao and Liu, Hongwei and Liu, Songlin},
	year         = 2025,
	journal      = {IEEE Transactions on Geoscience and Remote Sensing},
	volume       = 63,
	pages        = {1--14}
}

@article{an_2025,
	title        = {SAR Images Despeckling Using Subaperture Decomposition and Non-Local Low-Rank Tensor Approximation},
	author       = {An, Xinwei and Zeng, Hongcheng and Li, Zhaohong and Yang, Wei and Xiong, Wei and Wang, Yamin and Liu, Yanfang},
	year         = 2025,
	journal      = {Remote Sensing},
	publisher    = {MDPI},
	volume       = 17,
	number       = 15,
	pages        = 2716
}

@inproceedings{ristea_2022,
	title        = {Guided deep learning by subaperture decomposition: Ocean patterns from SAR imagery},
	author       = {Ristea, Nicolae-C{\u{a}}t{\u{a}}lin and Anghel, Andrei and Datcu, Mihai and Chapron, Bertrand},
	year         = 2022,
	booktitle    = {IGARSS 2022-2022 IEEE International Geoscience and Remote Sensing Symposium},
	pages        = {6825--6828},
	organization = {IEEE}
}

@article{ristea_2023,
	title        = {Guided unsupervised learning by subaperture decomposition for ocean SAR image retrieval},
	author       = {Ristea, Nicolae-C{\u{a}}t{\u{a}}lin and Anghel, Andrei and Datcu, Mihai and Chapron, Bertrand},
	year         = 2023,
	journal      = {IEEE Transactions on Geoscience and Remote Sensing},
	publisher    = {IEEE},
	volume       = 61,
	pages        = {1--11}
}

@article{iqbal_2025,
	title        = {Subaperture decomposition analysis for accurate ship detection and velocity estimation in synthetic aperture radar imagery},
	author       = {Iqbal, Muhammad Amjad and Anghel, Andrei and Datcu, Mihai},
	year         = 2025,
	journal      = {Remote Sensing Letters},
	publisher    = {Taylor \& Francis},
	volume       = 16,
	number       = 1,
	pages        = {55--65}
}

@article{wang_2023,
	title        = {Target recognition in SAR images using complex-valued network guided with sub-aperture decomposition},
	author       = {Wang, Ruonan and Wang, Zhaocheng and Chen, Yu and Kang, Hailong and Luo, Feng and Liu, Yingxi},
	year         = 2023,
	journal      = {Remote Sensing},
	publisher    = {MDPI},
	volume       = 15,
	number       = 16,
	pages        = 4031
}

@inproceedings{filipponi_2019,
	title        = {Sentinel-1 GRD preprocessing workflow},
	author       = {Filipponi, Federico},
	year         = 2019,
	booktitle    = {International Electronic Conference on Remote Sensing},
	pages        = 11,
	organization = {MDPI}
}

@article{deledalle_2012,
	title        = {How to Compare Noisy Patches? Patch Similarity Beyond Gaussian Noise},
	author       = {Charles-Alban Deledalle and Lo{\"i}c Denis and Florence Tupin},
	year         = 2012,
	month        = {Aug},
	day          = {01},
	journal      = {International Journal of Computer Vision},
	volume       = 99,
	number       = 1,
	pages        = {86--102},
	issn         = {1573-1405}
}

@misc{Snap,
	title        = {SNAP - ESA Sentinel Application Platform},
	author       = {European Space Agency}
}

@article{fan_2020,
	title        = {MA-Net: A Multi-Scale Attention Network for Liver and Tumor Segmentation},
	author       = {Fan, Tongle and Wang, Guanglei and Li, Yan and Wang, Hongrui},
	year         = 2020,
	journal      = {IEEE Access},
	volume       = 8,
	number       = {},
	pages        = {179656--179665}
}

@article{chamorro_2021,
	title        = {Fully convolutional recurrent networks for multidate crop recognition from multitemporal image sequences},
	author       = {Jorge Andres {Chamorro Martinez} and Laura Elena {Cu{\'e} La Rosa} and Raul Queiroz Feitosa and Ieda {Del{\'A}rco} Sanches and Patrick Nigri Happ},
	year         = 2021,
	journal      = {ISPRS Journal of Photogrammetry and Remote Sensing},
	volume       = 171,
	pages        = {188--201},
	issn         = {0924-2716}
}

@article{ko_2022,
	title        = {SAR Image Despeckling Using Continuous Attention Module},
	author       = {Ko, Jaekyun and Lee, Sanghwan},
	year         = 2022,
	journal      = {IEEE Journal of Selected Topics in Applied Earth Observations and Remote Sensing},
	volume       = 15,
	number       = {},
	pages        = {3--19}
}

@article{wang_2004,
	title        = {Image quality assessment: from error visibility to structural similarity},
	author       = {Wang, Zhou and Bovik, Alan C and Sheikh, Hamid R and Simoncelli, Eero P},
	year         = 2004,
	journal      = {IEEE transactions on image processing},
	publisher    = {IEEE},
	volume       = 13,
	number       = 4,
	pages        = {600--612}
}

@inproceedings{smith_2018,
	title        = {Super-convergence: very fast training of neural networks using large learning rates},
	author       = {Leslie N. Smith and Nicholay Topin},
	year         = 2018,
	booktitle    = {Defense + Commercial Sensing}
}

@inproceedings{kingma_2015,
	title        = {Adam: A Method for Stochastic Optimization},
	author       = {Kingma, Diederik P. and Ba, Jimmy},
	year         = 2015,
	booktitle    = {International Conference on Learning Representations (ICLR)}
}

@article{gomez_alanis_2020,
	title        = {A Kernel Density Estimation Based Loss Function and its Application to ASV-Spoofing Detection},
	author       = {Gomez-Alanis, Alejandro and Gonzalez-Lopez, Jose A. and Peinado, Antonio M.},
	year         = 2020,
	journal      = {IEEE Access},
	volume       = 8,
	number       = {},
	pages        = {108530--108543}
}

@article{massarelli_2026,
	title        = {Super-Resolution of Sentinel-2 Satellite Images: A Comparison of Different Interpolation Methods for Spatial Knowledge Extraction},
	author       = {Massarelli, Carmine},
	year         = 2026,
	journal      = {Machine Learning and Knowledge Extraction},
	volume       = 8,
	number       = 1,
	issn         = {2504-4990},
	article-number = 14
}
}

\end{document}